\documentclass[conference]{IEEEtran}

\usepackage{graphicx}
\usepackage{booktabs}
\usepackage{url}
\usepackage{array}
\usepackage{enumitem}
\newcommand{\PreserveBackslash}[1]{\let\temp=\\#1\let\\=\temp}
\newcolumntype{C}[1]{>{\PreserveBackslash\centering}p{#1}}
\newcolumntype{R}[1]{>{\PreserveBackslash\raggedleft}p{#1}}
\newcolumntype{L}[1]{>{\PreserveBackslash\raggedright}p{#1}}

\usepackage{colortbl}
\usepackage[table]{xcolor}

\usepackage{array}

\begin{document}

\title{Effect of Selection Format on LLM Performance

\author{
\IEEEauthorblockN{Yuchen Han\textsuperscript{1}, Yucheng Wu\textsuperscript{2}, Jeffrey Willard\textsuperscript{3}}
\IEEEauthorblockA{\textsuperscript{1}Washington University in St. Louis \textsuperscript{2}Nanjing University of Posts and Telecommunications, \textsuperscript{3}Boston University}
\IEEEauthorblockA{y.han@wustl.edu, p22000712@njupt.edu.cn, jeffwil@bu.edu}

}
}

\maketitle

\begin{abstract}
This paper investigates a critical aspect of large language model (LLM) performance: the optimal formatting of classification task options in prompts. Through an extensive experimental study, we compared two selection formats---bullet points and plain English---to determine their impact on model performance. Our findings suggest that presenting options via bullet points generally yields better results, although there are some exceptions. Furthermore, our research highlights the need for continued exploration of option formatting to drive further improvements in model performance.
\end{abstract}

\begin{IEEEkeywords}
natural language processing, prompt engineering, domain-specific tasks 
\end{IEEEkeywords}

\section{Introduction}

Large Language Models (LLMs) have reshaped how humans interact with technology and access information. Their ability to generate human-like text, understand complex queries, and perform a wide range of language-related tasks has made them very useful across various fields, such as clinical trials, education, and content generation. For example, AI-powered systems help streamline clinical documentation and provide symptom assessments through virtual health assistants. LLMs are used in tutoring platforms to offer personalized learning experiences. In creative fields, LLMs aid scriptwriting, generate game dialogue, and assist in content recommendations on streaming platforms. Besides that, AI methods are also commonly used in text detection~\cite{9744124,10043020,10172278,9747331,yang2022growinginstancemaskleaf,9900473}, traffic sign interpretation~\cite{10609794,yang2024signeyetrafficsigninterpretation}, traffic flow prediction and optimization~\cite{huang2024learningoptimalpathdnn,8955653,10851528}, financial time series forecasting~\cite{10851487}, network planning~\cite{zhao2017htn,zhao2017resource,liu2016hierarchical,qi2017hierarchical,wang2016review}, text understanding~\cite{rao2024machine}, quantum computing~\cite{li2024exploring}, financial and corporate analysis~\cite{yin2024predicting,xu2024energy,min2024financial}, industrial system understanding~\cite{diao2024ventilator,shen2024accurate,su2022mixed}, and medical application~\cite{wu2024multi,wang2024rpf}.

Prompt design is incredibly essential in shaping the quality, accuracy, and relevance of LLMs' outputs. A well-crafted prompt serves as the bridge between user intent and model response, guiding LLMs to better understand context, adhere to constraints, and deliver precise results tailored to specific tasks. Whether generating creative content, solving technical problems, or interpreting complex queries, the effectiveness of an LLM heavily depends on how it is prompted. As applications of LLMs continue to expand, prompt engineering becomes important to maximize their capabilities, enhancing user experience, and ensuring reliable performance in realistic scenarios.

Prompts have been extensively studied, with researchers examining various facets such as design, content, and context. However, one dimension remains shrouded in mystery: the selection format. This paper focuses on two selection formats for classification tasks - bullet point (BP) lists, which present LLMs with bullet points to choose from, and plain description (PD), which utilizes plain English to outline the available options.
Table~\ref{tab:selection_format_exp} illustrates the two selection formats. The left column presents sentiment options as three distinct bullet points: Negative, Neutral, and Positive. In contrast, the right column uses a sentence to prompt LLMs to determine the sentiment.

\begin{table}
    \centering
    \begin{tabular}{L{1.35in}|L{1.35in}}
        \toprule
        Bullet point (BP) 
        & Plain description (PD) \\
        \midrule
        \begin{minipage}{\linewidth}
            What is the sentiment of the sentence below? \\
            Dallas is fantastic!
            \begin{itemize}[nolistsep]
                \item Negative
                \item Neutral
                \item Positive
            \end{itemize}
        \end{minipage}
        & \begin{minipage}{\linewidth}
            What is the sentiment of the sentence below? \\
            Dallas is fantastic! \\
            Is it negative, neutral, or positive?
        \end{minipage} \\
        \bottomrule
        \addlinespace
    \end{tabular}
    \caption{Examples of two selection formats: Bullet point (BP) and Plain description (PD).}
    \label{tab:selection_format_exp}
\end{table}

To address the comparative impacts of two-format prompts, we choose ten pioneering research as base works categorized in Table~\ref{table:results}, 
which are beneficial domains to the regional advancements in terms of economics, health, and technologies. Our main contributions are as follows.
\begin{itemize}[noitemsep]
    \item We have formally proposed the research question: Does using bullet points outperform plain description in prompts for LLMs?
    \item Through a comprehensive experimental study spanning nine domain-specific tasks, our findings indicate that bullet points generally yield better results than plain descriptions.
    \item Building on our experiment results, we propose potential directions for future research to improve LLM performance further.
\end{itemize}

\section{Background}

\subsection{Large Language Model}

With the introduction of AlexNet~\cite{NIPS2012_c399862d}, we entered a new era of deep learning, marked by significant progress across various fields, such as image generation~\cite{yang2024mmoigmulticlassmultiscaleobject}, stereo matchin~\cite{Zhang2023EHSSAE}, map construction~\cite{zhang2024mapexpertonlinehdmap}, and AI agent~\cite{jia2024decentralized,jia2025embodied}. In NLP, the Transformer architecture~\cite{NIPS2017_3f5ee243} has revolutionized the entire field, serving as the backbone for modern language models (especially for large models). The two core components of the Transformer, the encoder and decoder, have laid the foundation for much of the work in model pretraining. BERT~\cite{devlin-etal-2019-bert} is a prominent example of encoder-based pretraining, inspiring numerous variants aimed at improving the pretraining paradigm, such as RoBERTa~\cite{DBLP:journals/corr/abs-1907-11692} and DistillBERT~\cite{Sanh2019DistilBERTAD}, or addressing domain-specific tasks, including SciBERT~\cite{beltagy-etal-2019-scibert} and ClinicalBERT~\cite{clinicalbert}.

Concurrently, another branch of pretraining research has focused on the decoder component of the Transformer architecture. Notable examples of this approach include the GPT-series models~\cite{Radford2018ImprovingLU,Radford2019LanguageMA,NEURIPS2020_1457c0d6} and the LLaMA-series models~\cite{touvron2023llamaopenefficientfoundation,touvron2023llama2openfoundation,grattafiori2024llama3herdmodels}. These models utilize the decoder as their primary backbone architecture, diverging from the encoder-based approach. The emergence of ChatGPT\footnote{chatgpt.com} sparked widespread interest in LLMs, leading researchers to shift their attention towards evaluating the effectiveness of these models.

Despite the increasing interest in LLMs, a significant knowledge gap remains: the impact of prompt length on the performance of LLaMA models in domain-specific tasks. This oversight is particularly significant, given that previous studies have demonstrated the critical role of prompts in unlocking LLMs' language understanding and task execution capabilities~\cite{10458651,10.1145/3655497.3655500,white2023promptpatterncatalogenhance,10698605,xiao-etal-2024-analyzing}. By addressing this gap, this paper aims to provide valuable insights into prompt design for domain-specific tasks, offering practical guidance for academics and industry professionals seeking to develop effective prompt strategies for their applications.

\subsection{Prompt Engineering}

Existing research on prompt engineering has primarily focused on enhancing the reasoning and logical capabilities of language models, with Chain-of-Thought (CoT) prompting~\cite{10.5555/3600270.3602070,zhang2022automaticchainthoughtprompting} being a pioneering approach that systematically investigates step-by-step reasoning. Building upon CoT, several subsequent advancements have been proposed to further improve LLMs' reasoning abilities, including self-consistency~\cite{wang2023selfconsistencyimproveschainthought}, which introduces mechanisms to improve reliability in reasoning paths, and Logical CoT prompting~\cite{zhao-etal-2024-enhancing-zero}, which refines logical reasoning within prompts. 
Additionally, Chain-of-Symbols (CoS) prompting~\cite{hu2024chainofsymbolpromptingelicitsplanning} explores symbolic representations to enhance task-solving processes, 
Tree-of-Thoughts prompting~\cite{10.5555/3666122.3666639} introduces hierarchical reasoning structures, 
and Chain-of-Tables prompting~\cite{wang2024chainoftableevolvingtablesreasoning} adapts tabular representations for specific reasoning tasks. 
Despite significant progress in enhancing LLMs' ability to tackle intricate reasoning and logic-based challenges, a crucial aspect remains unexplored: the impact of prompt formatting on the performance of LLMs in specialized domains.

\section{Experiments}

\subsection{LLM Evaluation Tasks}

To comprehensively study the prompt formats' reflection on LLMs' performance, we select ten novel types of research as a testbed, which represent diverse practices that LLMs are involved in. Specifically, 
\textbf{MPU (Monetary Policy Understanding)}~\cite{10871796} identifies the climate of monetary policy statements such as hawkish, dovish, or neutral, 
\textbf{UI (User Intent)}~\cite{10.1145/3637528.3671647} Categorizes users' intentions based on their statements.
\textbf{QIC (Query Intent Classification)}~\cite{javadi2024llmbasedweaksupervisionframework} discovers the purpose behind user queries, such as seeking information, navigation, or specific documents.
\textbf{SD (Sarcasm Detection)}~\cite{10.1145/3696271.3696294} detects whether a statement contains sarcasm.
\textbf{EI (Emotion Identification)}~\cite{10.1145/3696271.3696292} aims to better interpret emotional tone and context by classifying emotions as anger, joy, sadness, surprise, fear, and love. 
\textbf{FSA (Financial Sentiment Analysis)}~\cite{10.1145/3696271.3696299} analyzes financial texts to determine if the sentiment is positive, negative, or neutral.
\textbf{CSD (Communication System Delay)}~\cite{zhang2024empowering} examines communication system performance, e.g., the indicator of end-to-end delay, significant to the quality and effectiveness of communications.
\textbf{DD (Disease Detection)}~\cite{wei2024enhancingdiseasedetectionradiology} finds deviating items based on radiology reports mapped to ICD-10 codes.
\textbf{BioQA (Biomedical Question Answering}~\cite{li2024benchmarkingretrievalaugmentedlargelanguage} answers domain-specific questions in the biomedical field.
\textbf{PTI (Political Truth Identification)}~\cite{chatrath2024factfictionllmsreliable} identifies the accuracy of political statements.

\begin{table*}[ht]
    \centering
    \small
    \renewcommand\arraystretch{1.1}{
    \begin{tabular}{L{0.35in}|@{ }c@{ }|@{ }c|@{ }c|@{ }c@{ }|@{ }c|@{ }c|@{ }c@{ }|@{ }c|@{ }c}
        \toprule
        & \multicolumn{3}{@{ }c|@{ }}{Precision}             & \multicolumn{3}{@{ }c|@{ }}{Recall}                & \multicolumn{3}{@{ }c@{ }}{F1 Score}    \\
        \midrule
            & Base & BP  & PD   & Base & BP  & PD   & Base & BP  & PD   \\
        \midrule
            
            
        
        MPU & 0.56 
            & 0.59 (\colorbox{green!75}{\makebox[0.25in][c]{$\uparrow$ 0.03}}) 
            & 0.53 (\colorbox{red!60}{\makebox[0.25in][c]{$\downarrow$ 0.03}})
            
            & 0.55 
            & 0.57 (\colorbox{green!25}{\makebox[0.25in][c]{$\uparrow$ 0.02}}) 
            & 0.50 (\colorbox{red!60}{\makebox[0.25in][c]{$\downarrow$ 0.05}}) 
            
            & 0.54 
            & 0.58 (\colorbox{green!75}{\makebox[0.25in][c]{$\uparrow$ 0.04}}) 
            & 0.51 (\colorbox{red!60}{\makebox[0.25in][c]{$\downarrow$ 0.03}})  \\
    
        \midrule
        
        UI  & 0.52 
            & 0.54 (\colorbox{green!25}{\makebox[0.25in][c]{$\uparrow$ 0.02}})
            & 0.51 (\colorbox{red!25}{\makebox[0.22in][c]{$\downarrow$ 0.01}})  
            & 0.65 
            & 0.62 (\colorbox{red!60}{\makebox[0.25in][c]{$\downarrow$ 0.03}}) 
            & 0.56 (\colorbox{red!60}{\makebox[0.25in][c]{$\downarrow$ 0.09}})   
            & 0.62 
            & 0.58 (\colorbox{red!60}{\makebox[0.25in][c]{$\downarrow$ 0.04}})
            & 0.53 (\colorbox{red!60}{\makebox[0.25in][c]{$\downarrow$ 0.09}})  \\
        \midrule
        QIC & 0.50 
            & 0.56 (\colorbox{green!75}{\makebox[0.25in][c]{$\uparrow$ 0.06}})
            & 0.48 (\colorbox{red!25}{\makebox[0.25in][c]{$\downarrow$ 0.02}})  
            & 0.30 
            & 0.37 (\colorbox{green!75}{\makebox[0.25in][c]{$\uparrow$ 0.07}}) 
            & 0.25 (\colorbox{red!60}{\makebox[0.25in][c]{$\downarrow$ 0.05}})   
            & 0.40
            & 0.48 (\colorbox{green!75}{\makebox[0.25in][c]{$\uparrow$ 0.08}})
            & 0.37 (\colorbox{red!60}{\makebox[0.25in][c]{$\downarrow$ 0.03}})  \\
         \midrule
         
        SD  & 0.67 
            & 0.71 (\colorbox{green!75}{\makebox[0.25in][c]{$\uparrow$ 0.04}})
            & 0.65 (\colorbox{red!25}{\makebox[0.25in][c]{$\downarrow$ 0.02}})  
            & 0.66 
            & 0.70 (\colorbox{green!75}{\makebox[0.25in][c]{$\uparrow$ 0.03}}) 
            & 0.62 (\colorbox{red!60}{\makebox[0.25in][c]{$\downarrow$ 0.04}})   
            & 0.67
            & 0.70 (\colorbox{green!75}{\makebox[0.25in][c]{$\uparrow$ 0.03}})
            & 0.63 (\colorbox{red!60}{\makebox[0.25in][c]{$\downarrow$ 0.04}})  \\
         \midrule
         
        EI  & 0.50 
            & 0.51 (\colorbox{green!75}{\makebox[0.25in][c]{$\uparrow$ 0.06}})
            & 0.50
            & 0.49 
            & 0.50 (\colorbox{green!25}{\makebox[0.25in][c]{$\uparrow$0.01}}) 
            & 0.45 (\colorbox{red!60}{\makebox[0.25in][c]{$\downarrow$ 0.04}})   
            & 0.49 
            & 0.50 (\colorbox{green!25}{\makebox[0.25in][c]{$\uparrow$ 0.01}})
            & 0.47 (\colorbox{red!25}{\makebox[0.25in][c]{$\downarrow$ 0.02}})  \\
         \midrule
         
        FSA & 0.71 
            & 0.73 (\colorbox{green!25}{\makebox[0.25in][c]{$\uparrow$ 0.02}})
            & 0.70 (\colorbox{red!25}{\makebox[0.25in][c]{$\downarrow$ 0.01}})  
            & 0.75 
            & 0.76 (\colorbox{green!25}{\makebox[0.25in][c]{$\uparrow$0.01}}) 
            & 0.76 (\colorbox{green!25}{\makebox[0.25in][c]{$\uparrow$ 0.01}})   
            & 0.72 
            & 0.75 (\colorbox{green!75}{\makebox[0.25in][c]{$\uparrow$ 0.03}})
            & 0.72  \\
         \midrule
    
        CSD & 0.45 
            & 0.45
            & 0.44 (\colorbox{red!25}{\makebox[0.22in][c]{$\downarrow$ 0.01}})  
            & 0.47 
            & 0.49 (\colorbox{green!25}{\makebox[0.25in][c]{$\uparrow$ 0.02}}) 
            & 0.46 (\colorbox{red!25}{\makebox[0.22in][c]{$\downarrow$ 0.01}})  
            & 0.46 
            & 0.47 (\colorbox{green!25}{\makebox[0.25in][c]{$\uparrow$ 0.01}})
            & 0.45 (\colorbox{green!25}{\makebox[0.25in][c]{$\uparrow$ 0.01}})  \\
         \midrule
         
        DD & 0.83 
            & 0.85 (\colorbox{green!25}{\makebox[0.25in][c]{$\uparrow$ 0.02}})
            & 0.83  
            & 0.88 
            & 0.89 (\colorbox{green!25}{\makebox[0.25in][c]{$\uparrow$ 0.01}}) 
            & 0.87 (\colorbox{red!25}{\makebox[0.25in][c]{$\downarrow$ 0.02}})   
            & 0.87 
            & 0.86 (\colorbox{green!25}{\makebox[0.25in][c]{$\uparrow$ 0.01}})
            & 0.86 (\colorbox{red!25}{\makebox[0.25in][c]{$\downarrow$ 0.01}})  \\
         \midrule
        BioQA & 0.82 
            & 0.85 (\colorbox{green!75}{\makebox[0.25in][c]{$\uparrow$ 0.03}})
            & 0.83 (\colorbox{green!25}{\makebox[0.25in][c]{$\uparrow$ 0.01}})  
            & 0.81 
            & 0.82 (\colorbox{green!25}{\makebox[0.25in][c]{$\uparrow$ 0.01}}) 
            & 0.82 (\colorbox{green!25}{\makebox[0.25in][c]{$\uparrow$ 0.01}})   
            & 0.81 
            & 0.83 (\colorbox{green!25}{\makebox[0.25in][c]{$\uparrow$ 0.02}})
            & 0.81  \\
         \midrule
        PTI & 0.78 
            & 0.80 (\colorbox{green!25}{\makebox[0.25in][c]{$\uparrow$ 0.02}})
            & 0.79 (\colorbox{green!25}{\makebox[0.25in][c]{$\uparrow$ 0.01}})  
            & 0.80 
            & 0.82 (\colorbox{green!25}{\makebox[0.25in][c]{$\uparrow$ 0.02}}) 
            & 0.80   
            & 0.79 
            & 0.81 (\colorbox{green!25}{\makebox[0.25in][c]{$\uparrow$ 0.02}})
            & 0.79  \\
         
        \bottomrule
        \addlinespace
    \end{tabular}}
    \caption{Experimental results with different selection formats. 
    }
    \label{table:results}
\end{table*}

\subsection{Experiment Design}

In our experiment, each group consists of three settings: the prompt used in the original paper (e.g., MPU\textsubscript{base}), the prompt using bullet points (e.g., MPU\textsubscript{BP}), and the prompt using plain description (e.g., MPU\textsubscript{PD}). Note that the default prompts used in previous work on these domain-specific tasks may have been formatted in BP, PD, or other formats\footnote{They could also use mixed selection format.}. In our study, we converted them to strictly BP or PD to investigate how the selection format impacts LLM performance. Additionally, we did not include any examples in the prompt to isolate the effect of the selection format, ensuring that only this component was taken into account.

\subsection{Evaluation Metrics}

We use the weighted average precision, recall, and F1-score for each experiment as the evaluation metrics, where weight is determined by the number of instances for each class. We denote true positive, false positive, true negative, and false negative as \emph{TP, FP, TN}, and \emph{FN}. Precision reflects the proportion of correctly predicted positive instances (True Positives, TP) out of all predicted positive instances, including False Positives (FP). Recall indicates the proportion of actual positive instances correctly identified by the model, accounting for both True Positives (TP) and False Negatives (FN). F1 score Combines precision and recall into a single metric by calculating their harmonic mean\footnote{Metrics Equations: $precision = \frac{TP}{TP+FP}, 
recall = \frac{TP}{TP+FN}, 
F1 = \frac{2 * precision * recall}{precision + recall}$.}

\section{Results}
Table~\ref{table:results} presents performance comparisons across different categorized tasks for Precision, Recall, and F1-score metrics using three prompt formats: Base, BP, and PD. Key observations are as follows.

BP format consistently improves precision over the base format across most categories, with notable increases for MPU (+0.03), QIC (+0.06), SD (+0.04), and BioQA (+0.03).
PD format shows a drop in precision for many categories compared to the base and BP formats, such as MPU (-0.03), QIC (-0.02), and SD (-0.02).

Regarding recall, BP format shows more effectiveness over base format with QIC (+0.07) and SD (+0.03) while maintaining stable results in other categories.
PD format significantly impacts recall for multiple categories, including MPU (-0.05), UI (-0.09), QIC (-0.05), SD (-0.04), and EI (-0.04).

Finally, BP format improves F1 in categories like MPU (+0.04), QIC (+0.08), SD (+0.03), and FSA (+0.03).
PD format decreases F1 for many categories, with the same declines as its impacts on recall.

The BP format generally improves performance across most tasks, with notable gains in recall and F1 score. This suggests that BP format is more effective at handling both false positives and false negatives compared to the base format. The PD method frequently underperforms, leading to reductions in precision, recall, and F1 scores. This implies potential issues with PD format's ability to generalize or maintain prediction accuracy across diverse datasets.

\section{Conclusions}

In this paper, we conducted extensive experiments to evaluate LLM performance on domain-specific tasks by comparing two selection formats. Our findings reveal that presenting options as bullet points generally outperforms using plain descriptions. We attribute this difference to the nature of the corpus for LLM pretraining. Specifically, if LLMs are pretrained on datasets rich in bullet-point formats, they are better equipped to process and follow such structures than plain English sentences.

Our future research will analyze whether the formats in pretraining or finetuning corpora influence model performance in downstream tasks. For instance, prior work by~\cite{xiao-etal-2024-analyzing,xiao-etal-2023-context,xiao-blanco-2022-people} demonstrated that finetuning on the corpus with spatial annotations yielded strong results in spatial information extraction. At the same time, \cite{meyer2024comparisonllmfinetuningmethods} achieved success in travel annotation through fine-tuning on a domain-specific corpus. On the other hand, although LLMs become more and more popular in many areas, such as recommendation system~\cite{yu2025applicationlargelanguagemodels} and fake news detection~\cite{yi2025challengesinnovationsllmpoweredfake,xu2025hybrid}, privacy risks started to gain substantial attraction~\cite{10851615}. To address privacy issues, we also aim to analyze how the differential privacy can be guaranteed when leveraging LLMs to tackle domain-specific tasks.

\bibliography{cm2}
\bibliographystyle{IEEEtran}

\end{document}